\documentclass[letterpaper, 10 pt, conference]{ieeeconf}  

\IEEEoverridecommandlockouts                              

\usepackage{amsmath}
\usepackage{graphicx}
\usepackage{amsmath,amssymb}
\usepackage{scalerel}
\usepackage[breaklinks]{hyperref}

\let\bigopsize\bigoplus
\def\bigoplus{{\scalerel*{\boldsymbol\oplus}{\bigopsize}}}

\makeatletter 
\let\proof\@undefined
\let\endproof\@undefined
\let\NAT@parse\undefined
\makeatother
\usepackage[numbers, sort, compress]{natbib}

\usepackage{breakurl}
\usepackage{xcolor}

\usepackage{fancyhdr}
\fancypagestyle{withfooter}{
  
  \fancyhead[L]{}
  \fancyhead[R]{}
  \fancyfoot[C]{\footnotesize Presented at the 2025 IEEE ICRA Workshop on Field Robotics}
}
\title{\LARGE \bf
Factor-Graph-Based Passive Acoustic Navigation for Decentralized Cooperative Localization Using Bearing Elevation Depth Difference
}

\author{Kalliyan Velasco, Timothy~W.~McLain,~and Joshua G. Mangelson%
  \thanks{This work was funded by the Naval Sea Systems Command (NAVSEA), Naval Surface Warfare Center - Panama City Division (NSWC-PCD) under the Naval Engineering Education Consortium (NEEC) Grant Program under award number N00174-23-1-0005. This work was also partially funded by the Center for Autonomous Air Mobility and Sensing (CAAMS), a National Science Foundation Industry-University Cooperative Research Center (IUCRC) under NSF award number 2139551, along with significant contributions from CAAMS industry members.}
  \noindent
  \thanks{K.~Velasco, T.~McLain, and J.~Mangelson are at Brigham Young University. They can be reached at \texttt{\{kalliyan, mclain, mangelson\}@byu.edu}.}  %
}

\begin{document}

\maketitle
\thispagestyle{withfooter}
\pagestyle{withfooter}

\begin{abstract}

Accurate and scalable underwater multi-agent localization remains a critical challenge due to the constraints of underwater communication. In this work, we propose a multi-agent localization framework using a factor-graph representation that incorporates bearing, elevation, and depth difference (BEDD). Our method leverages inverted ultra-short baseline (inverted-USBL) derived azimuth and elevation measurements from incoming acoustic signals and relative depth measurements to enable cooperative localization for a multi-robot team of autonomous underwater vehicles (AUVs). We validate our approach in the HoloOcean underwater simulator with a fleet of AUVs, demonstrating improved localization accuracy compared to dead reckoning. Additionally, we investigate the impact of azimuth and elevation measurement outliers, highlighting the need for robust outlier rejection techniques for acoustic signals.

\end{abstract}

\section{Introduction}
Localization in underwater environments presents significant challenges due to the absence of GPS and the feature-poor nature of the terrain. Although terrain-aided navigation (TAN) can help mitigate drift by matching observed bathymetry to known maps, it is ineffective in regions where bathymetry remains constant, leading to localization failures. Multi-agent localization can address these challenges by expanding the search space and sharing information to reduce uncertainty and decrease search time.

A promising approach to underwater localization is bearing, elevation, and depth-difference (BEDD) passive inverted acoustic navigation~\cite{sekimoriBEDD2021}. While prior studies have demonstrated the BEDD technique using a single surface vessel and an underwater vehicle~\cite{sekimoriBEDD2024}, our work seeks to extend it to decentralized localization for a fully underwater fleet. Additionally, instead of implementing BEDD using a particle filter, we present a factor-graph framework for multi-agent BEDD. Lastly, past implementations managed acoustic-modem outliers by rejecting elevation measurements shallower than 40~degrees. As we develop our method, we aim to define a bearing measurement consistency metric for outlier rejection using consistent set maximization~\cite{mangelsonPCM}, allowing more data to be utilized rather than discarded.

This paper outlines our multi-agent factor-graph-based localization approach implementing BEDD. Our initial efforts have demonstrated our method in the HoloOcean~\cite{potokarHoloOcean2022} underwater simulator. 

\begin{figure}
  \centering
  \includegraphics[width=0.47\textwidth]{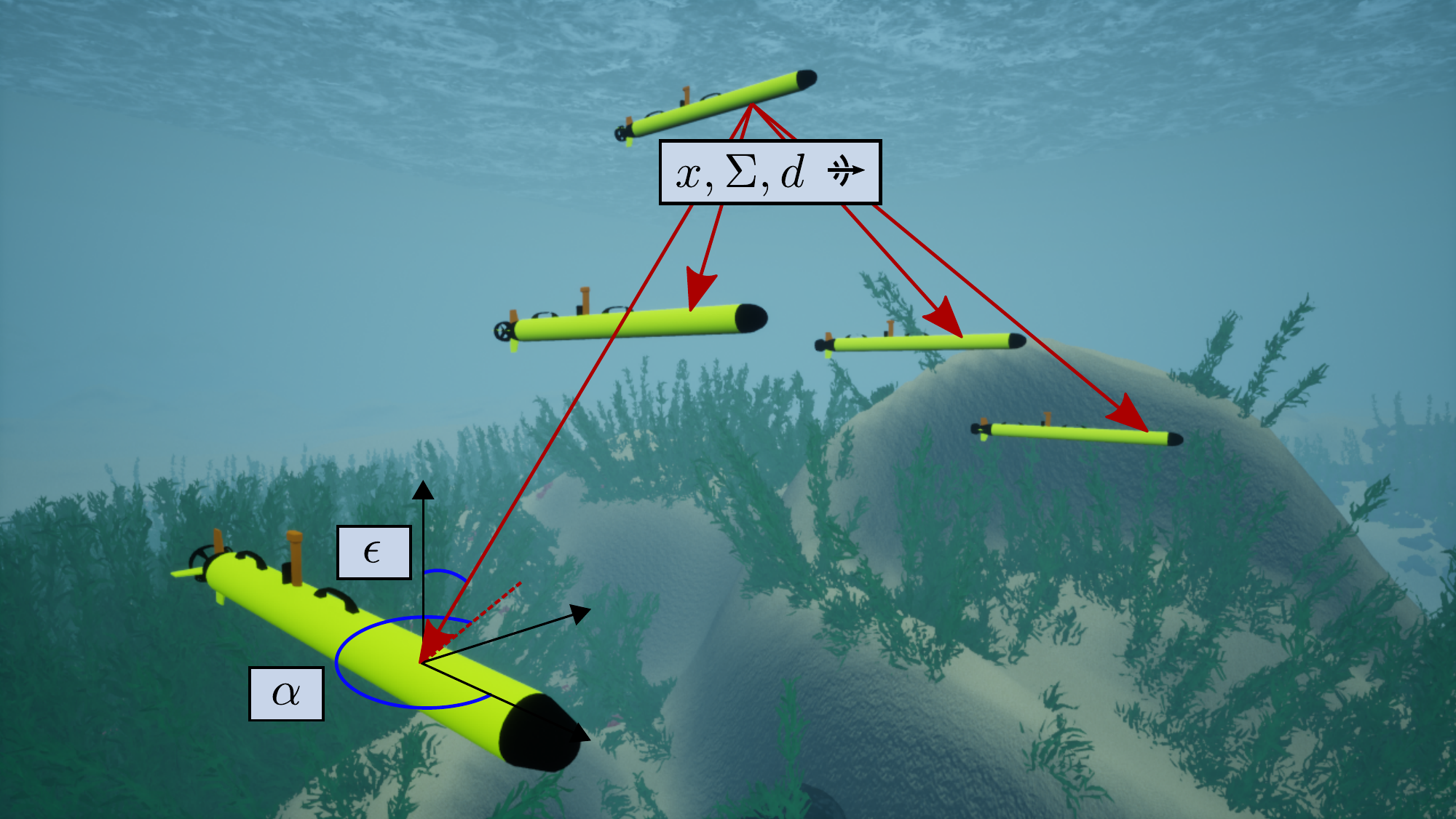}
  \caption{An example of multi-agent localization in the HoloOcean underwater simulator. The transmitting agent sends the measured depth $(d)$, composed odometry $(\mathbf{x})$, and marginalized covariance $(\Sigma)$ via an acoustic signal (red arrows). All neighboring agents passively receive the message and compute the azimuth $(\alpha)$ and elevation $(\epsilon)$ of the acoustic signal onboard. The received measurements are integrated into the multi-agent factor graph of each respective agent.}
\end{figure}

\section{Related Work}

\subsection{Decentralized Underwater Acoustic Positioning}

Ultra-short baseline localization (USBL) systems are a relatively common technique for tracking underwater assets. USBL typically operates by leveraging a surface-expressed agent (such as a surface vessel) with a transducer to transmit requests and a hydrophone array. On a predetermined cycle, the surface vessel in turn requests a response from each underwater agent, which then replies via an on-board transponder~\cite{fallonRTR2010}. The surface vessel determines the range to a target agent using the speed of sound underwater and the two-way time-of-flight (TW-ToF) of the acoustic signal, while azimuth and elevation are estimated using the transceiver’s phased hydrophone array.
 
This typical use-model is inherently centralized and risks system failure if the surface vessel malfunctions (in addition to requiring a surface expressed agent). Additionally, a centralized model  limits the distance underwater agents can operate from the central-node (and potentially the surface itself). As a result, several methods have been developed for decentralized localization using acoustic communication. 

\subsubsection{Two-way communication} 
In decentralized multi-agent localization using TW-ToF, each agent transmits a request to a neighbor and awaits a response to compute range~\cite{matsudaTWTF2021}. Since each agent must exchange signals with every neighbor, scalability is limited~\cite{rypkemaOWTT2017}. To mitigate this, \citet{gaoTWTT2014} proposed a scheduling method where all neighbors respond sequentially after a single broadcast request. However, the need for a response still limits scalability.

\subsubsection{One-way communication with synchronized clocks}
One-way time-of-flight (OW-ToF) improves scalability over TW-ToF by using synchronized atomic clocks~\cite{rypkemaOWTT2017, jakubaOWTT2015, wangDiscOWTT2022}. A transmitting agent broadcasts the time of launch (TOL), enabling neighbors to compute range without responding. However, underwater synchronization challenges make it difficult to add new agents to a fleet after deployment. This limits the applicability of this technique to long-term and large-scale multi-agent operations.

\begin{figure}
  \centering
  \includegraphics[width=0.47\textwidth]{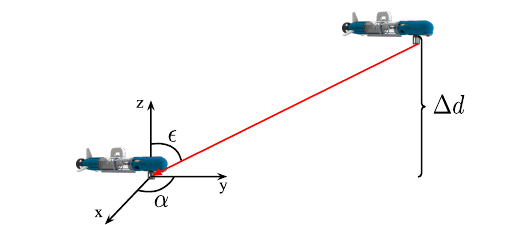}
  \caption {BEDD triangle geometry using azimuth $\alpha$, elevation $\epsilon$, and depth difference $\Delta d$.}
  \label{geometry}
\end{figure}

\subsubsection{One-way BEDD and inverted-USBL}
An alternative scalable positioning approach that does not rely on clock synchronization uses bearing, elevation, and depth difference (BEDD)~\cite{sekimoriBEDD2024}. In this method, every agent is equipped with a USBL hydrophone array (as opposed to a single centralized agent). A transmitting agent broadcasts measured depth from an onboard pressure sensor. The acoustic signal is passively received by all neighbors and used to determine the depth difference, azimuth and elevation. Using basic triangulation, as illustrated in Fig. \ref{geometry}, the receiving agents estimate the position of the transmitting agent. Some limitations to BEDD include restrictions to non-planar fleet configurations and sensitivity to acoustic bearing noise.

\subsection{Acoustic Communication Outlier Rejection}

A key challenge in acoustic localization is mitigating the impact of multipath effects. Previous approaches to handling multipath include elevation thresholding, where all elevation measurements considered "too shallow" are rejected~\cite{sekimoriBEDD2021}, Mahalanobis-distance filtering~\cite{vaganayMahalRange1996}, which discards range measurements that significantly deviate from predicted beacon travel times, and group-k consistency maximization~\cite{forsgrenGCKM2024}, which identifies the maximum clique of consistent range measurements. While the last two methods have been effective for range-based acoustic navigation, outlier rejection for bearing-only measurements has not been fully explored in this context. Developing robust outlier rejection techniques for bearing-based localization remains an open challenge and a crucial step towards improving BEDD-based multi-agent localization.

\subsection{Graph-Based Localization}

Previous implementations of BEDD localization have relied on particle-filter-based localization due to its ability to handle non-Gaussian models effectively~\cite{sekimoriBEDD2024}. However, improving the probability distribution representation requires increasing the number of tracked particles, making particle filters computationally expensive. In addition, previous particle-filter-based BEDD implementations use the relative transformations between agents only for estimating the position of the local AUV in a fleet, without tracking the states of neighboring agents.

In contrast, factor-graph-based multi-agent localization optimizes pose estimates for all agents over the joint probability distribution of process and measurement models~\cite{gtsam}. Previous works use factor graphs to address the challenges of acoustic communication dropout and bandwidth limitations by only broadcasting an approximation of a target agent's local odometry chain~\cite{paullBetweenFactor2015, wallsFactorComp2015}. Additionally, the use of anchor nodes for multi-agent systems~\cite{kimAnchorNode2010, brinkMultiFG2017} enables more relaxed assumptions on the initial poses of neighboring agents, and iteratively updates the predicted transformation to the global frame as part of the optimization problem. These works provide additional robustness needed for underwater multi-agent applications.


\section{Multi-agent Localization}

In our approach for underwater multi-agent localization, each agent maintains one multi-agent factor graph and one single-agent odometry pose graph. We refer to the agent generating a given graph as the \textit{computing agent}. During the mission, agents transmit acoustic messages in a predetermined cycle. When an agent broadcasts a message, it transmits its current composed pose estimate from its single-agent factor graph, marginalized covariance, and depth measurement. Upon receiving the acoustic signal, the receiving agent computes azimuth and elevation using the inverted-USBL hydrophone array. Rather than explicitly calculating the depth difference and relative pose between agents, the depth, elevation, and azimuth measurements are added directly as factors in the multi-agent graph. 

\subsection{Single-agent Odometry Pose Graph}
\begin{figure}
  \centering
  \includegraphics[width=0.47 \textwidth]{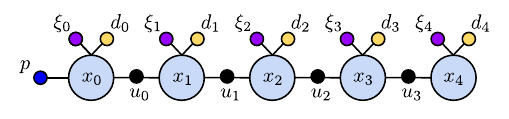}
  \caption{Single agent odometry pose graph with pose variables $x_i$, prior factor $p$, dead-reckoning factors $u_i$, orientation factors $\xi_i$, and depth factors $d_i$.}
  \label{single_fg}
\end{figure}

Factor graphs provide a visual representation of our multi-agent localization problem. An example of a single-agent pose graph is shown in Fig. \ref{single_fg}. Variable nodes correspond to parameters to be estimated and factor nodes encode constraints, or cost terms, to be optimized. Typically, each factor corresponds to a measurement (or piece of information) relating to the state of the robot at a specific point in time. Each factor encodes a residual cost term that measures the norm of the difference between the predicted measurement and the actual observed measurement. The factor graph encodes the conditional independence of the factors, given the state. The joint probability distribution over states and measurements is used to formulate a maximum a posteriori (MAP) estimation
problem, which can be solved with sparse non-linear optimization techniques~\cite{isam}. 

Assuming conditional independence among factors and Gaussian additive noise, the MAP estimation problem simplifies to nonlinear least-squares optimization, where each factor cost term becomes the squared Mahalanobis norm of the error between the predicted measurement (via evaluation of a measurement model at the current state estimate) and the observed values (as derived from on-board sensors or prior information).  In the following, we use $\|\mathbf{e}\|_{\Sigma}^{2}=\mathbf{e}^{T}\Sigma^{-1}\mathbf{e}$ to denote the squared Mahalanobis norm of the vector $\mathbf{e}$.  
%

Our full cooperative localization algorithm (as outlined in Section \ref{sec:multi-graph}) depends on the ability of each local agent to estimate its own local odometry. We formulate this local odometry estimation problem as the factor graph shown in Fig. \ref{single_fg}. The goal of the single-robot odometry estimation problem is to find the best estimate for the full trajectory of the agent by minimizing the error in observed dead-reckoning, orientation, and depth measurements. The corresponding non-linear optimization problem is expressed in \eqref{single_cost_funcs} as:

{\small
\begin{align}
    \mathrm{X}_{t}^{*} &= \underset {\mathrm{X}}{\text{arg min}}  
    \Bigg\{
        \underbrace{ \left\|P(\mathbf{x}_0,\mathbf{p})\right\| ^2_{\Sigma_{P}}}_{\text{Pose Prior}}
        + \sum_{i=0}^{M-1}  
            \Bigg(
            \underbrace{\left\|G(\mathbf{x}_{i-1},\mathbf{x}_{i}, \mathbf{u}_{i})\right\|_{\Sigma_{G_i}}^{2} }_{\text{Dead-reckoning}} 
            \Bigg)
            \nonumber \\
            & \quad
             + \sum_{i=0}^{M}  
            \Bigg(
            \underbrace{\left\|\Xi({\bf x}_{i}, \boldsymbol{\xi}_i)\right\|_{\Sigma_{\Xi_i}}^{2} }_{\text{Orientation}}
            + \underbrace{\left\|D({\bf x}_{i}
            ,d_i)\right\|_{\Sigma_{D_i}}^{2}}_{\text{Depth}}
            \Bigg)
    \Bigg\} 
    \label{single_cost_funcs}
\end{align}
}

\noindent
where each error function and variable will be defined in the corresponding factor sections.

For all but one cost term in this optimization problem, the error can be computed in the Lie group space, as rigid body transformations are naturally represented in $\mathrm{SE}(3)$, $\mathrm{SO}(3)$, and $\mathrm{SO}(2)$. The error in the group space can be calculated using
\begin{align}
    E(\hat M, M) = \mathrm{ln}(\hat{M}^{-1}M)^\vee ,
    \label{error_group}
\end{align}
where $E$ is the error function, $\hat M$ and $M$ are matrix elements of the Lie group and denote the predicted and measured values respectively, $\mathrm{ln}$ is the logarithm map for converting Lie group elements to the Lie algebra, and $\vee$ extracts the vector representation in Euclidean space as demonstrated by \cite{barfootLieDiff2017}. Moving forward, the error for all cost terms except for the depth cost term will be calculated in the group space.

Let the homogeneous transformation $\mathbf{x}_i \in \mathrm{SE}(3)$ denote the robot's pose at time step $i$ and let $\mathrm{X} = \{ \mathbf{x}_i \}_0^{M}$ be the set of all robot poses for all time steps from $0$ to $M$. With definition \eqref{error_group} in mind, factors in the proposed single-agent odometry pose graph are defined as follows:

\subsubsection{Pose prior} \label{sec:pose_prior}
A unary factor is a factor that depends only on one variable node. The unary prior factor $\mathbf{p} \in \mathrm{SE}(3)$ anchors the optimization problem by constraining the initial pose $\mathbf{x}_0$. Without this prior, the optimization problem would be insufficiently constrained and multiple possible solutions would exist. In our implementation, we choose the prior to be the local origin of the agent. The pose error is computed in the $\mathrm{SE}(3)$ group space using $P(\mathbf{x}_0,\mathbf{p}) = E(\hat{\mathbf{p}}, \mathbf{p})$, where $\hat{\mathbf{p}}=\mathbf{x}_0$.

\subsubsection{Dead-reckoning} \label{sec:dead_reck_single}
The dead-reckoning factor constrains motion from one time step to the next. The measurement $\mathbf{u}_{i} \in \mathrm{SE}(3)$ is retrieved from sensors such as an inertial measurement unit (IMU) or Doppler velocity log (DVL) on board the AUV. The predicted relative transformation $\hat{\mathbf{u}}_i \in \mathrm{SE}(3)$ from the previous pose $\mathbf{x}_{i-1}$ to the next pose $\mathbf{x}_{i}$ is computed using 
\begin{align}
    \hat{\mathbf{u}}_i &= \mathbf{x}^{-1}_{i-1}\mathbf{x}_{i}.
    \label{odom}
\end{align}
The error is then computed in the $\mathrm{SE}(3)$ group space using $G(\mathbf{x}_{i-1},\mathbf{x}_{i}, \mathbf{u}_{i})=E(\hat{\mathbf{u}}_{i},\mathbf{u}_{i})$.

\subsubsection{Orientation} \label{sec:orientation}
The unary orientation factor enforces constraints on the roll ($\phi$), pitch ($\theta$), and yaw ($\psi$) of the agent for each pose, based on measurements from the IMU. The measured orientation $\boldsymbol{\xi}_i \in \mathrm{SO}(3)$ is represented as a rotation matrix
{\small
\begin{align}
    \boldsymbol{\xi}_i =& R_z(\psi)R_y(\theta)R_x(\phi) \nonumber \\
    =& 
    \begin{bmatrix}
c_\psi & -s_\psi & 0 \\
s_\psi & c_\psi & 0 \\
0 & 0 & 1
\end{bmatrix}
\begin{bmatrix}
c_\theta & 0 & s_\theta \\
0 & 1 & 0 \\
-s_\theta & 0 & c_\theta
\end{bmatrix}
\begin{bmatrix}
1 & 0 & 0 \\
0 & c_\phi & -s_\phi \\
0 & s_\phi & c_\phi
\end{bmatrix},
\label{orientation}  
\end{align}
}
where $s_x := \sin(x)$ and $c_x := \cos(x)$, and $R$ denotes a rotation about one axis. The predicted $\hat {\boldsymbol{\xi}}_i\in \mathrm{SO}(3)$ is extracted from the predicted pose $\mathbf{x}_{i}$. Then, the orientation error is calculated in the $\mathrm{SO}(3)$ group space using $\Xi({\bf x}_{i}, \boldsymbol{\xi}_i)=E(\hat{\boldsymbol{\xi}}_i,\boldsymbol{\xi}_i)$.

\subsubsection{Depth} \label{sec:depth}
The unary depth factor uses scalar pressure sensor measurements $d_i$ to constrain the position of the agent along the vertical axis at each pose. The predicted depth $\hat d_i$ is obtained by extracting the $z$ component from $\mathbf{x}_{i}$, then used to calculate the error $D({\bf x}_{i},d_i)=\hat d_i-d_i$.
\begin{figure}
  \centering
  \includegraphics[width=0.47\textwidth]{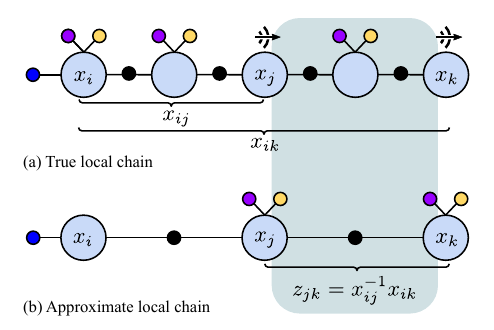}
  \caption{Factor decomposition using the origin state method (OSM). The transformation $x_{ij}$ denotes the relative transformation  from the origin $x_i$ to the last ToL $x_j$. The composition $x_{ik}$ denotes the relative transformation from the origin to the most recent ToL $x_k$. The two components can be decomposed to obtain the new odometry factor in the multi-agent graph $z_{jk}$. The shaded box highlights how the factors in the true local chain are summarized into the single factor $z_{jk}$ in the approximate local chain.}
  \label{OSM}
\end{figure}

\subsection{Origin State Method for Sharing Odometry}
\label{sec:origin-state}
During a mission, each agent periodically shares its odometry via a broadcast acoustic message. However, acoustic communications are severely limited both in capacity (on the order of 32 bytes per transmission) and time (as communication is time multiplexed between agents). Moreover, acoustic communications are highly lossy with transmissions often exceeding a 50~percent dropout rate. For robustness against dropout and low-bandwidth constraints associated with acoustic communication, we use the origin state method (OSM) proposed by \citet{wallsFactorComp2015} to share information between single and multi-agent factor graphs. 

First, we define a \textit{local chain} as the set of dead-reckoning and unary factors from a robot's single-agent odometry pose graph. Instead of broadcasting each factor from an agent's graph, \citet{wallsFactorComp2015} broadcast a decomposable summarized version of the local chain. Each agent simply transmits the odometry transformation between the local origin of the robot and the most recent pose at the time of launch (ToL). Fig. \ref{OSM}a shows an example of this process. At time $t=j$, the robot transmits the transformation between its local origin and its current pose estimate, $\mathbf{x}_{ij} \in \mathrm{SE}(3)$. Subsequently, at time $t=k$, the robot transmits the transformation between its local origin and its recently updated state estimate, $\mathbf{x}_{ik} \in \mathrm{SE}(3)$.

In addition to the transformation described above, the covariance and unary factors for depth and orientation at the time of transmission are also broadcast between agents. The covariance associated to $\mathbf{x}_{ik}$ ($\Sigma_{ik}$ at time step $k$ in Fig. \ref{OSM}a) can be calculated by extracting the marginal covariance from the optimization result and using methods proposed by~\cite{barfootUncertainty2014, mangelsonUncertainty2020}. Transmitted unary factors are not included in the local chain in order to avoid double counting in the multi-agent graph.

To recover an odometry factor from the transmitted summary from a remote agent, the receiving agent decomposes the most recently received link $\mathbf{x}_{ik}$ with the previously received link $\mathbf{x}_{ij}$ to reconstruct an approximated odometry transformation $\mathbf{z}_{jk}$ as follows: 
\begin{align}
    \mathbf{z}_{jk} =  \mathbf{x}_{ij}^{-1}\mathbf{x}_{ik}
    \label{between_operation}
\end{align}
This process is shown in Fig. \ref{OSM}b.

 Given transmitted summarizations $(\mathbf{x}_{ij}, \Sigma_{ij})$ and $(\mathbf{x}_{ik}, \Sigma_{ik})$, the covariance for the approximated factor, $\Sigma_{jk}$ can also be recovered as outlined in \cite{barfootUncertainty2014, mangelsonUncertainty2020}. 
The transformation $\mathbf{z}_{jk}$ and associated covariance $\Sigma_{jk}$ can then be used to form an approximated odometry measurement $\mathbf{z}_{i}^{r}$ within the multi-agent pose graph of the receiving agent as described in the next section. 

\begin{figure}
  \centering
  \includegraphics[width=0.47\textwidth]{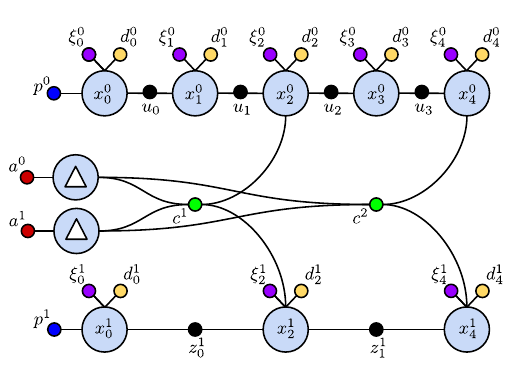}
  \caption{A multi-agent factor graph, with large circles representing variable nodes and small circles representing factors. The large blue circles indicate pose nodes, while the large blue circles with white triangles indicate the anchor nodes of each respective agent. For readability, connections from the anchor nodes to each orientation and depth factor were omitted, but the relationship is shown in \eqref{multi_cost_funcs}.}
  \label{multiagent_graph}
\end{figure}

\subsection{Multi-agent Relative Pose Graph}
\label{sec:multi-graph}

 The multi-agent graph (Fig. \ref{multiagent_graph}) allows the computing agent to bound its pose estimate uncertainty using information shared from neighboring agents. Let $R$ denote the number of robots, $M_r+1$ the number of pose variables for a given robot $r \in \{0, \dots, R-1\}$, and $N$ the number of inter-agent encounters seen by the computing agent. We refer to the computing agent as $r=0$. The factor graph is initialized with each agent starting at its own local origin. In addition to pose variables $\mathbf{x}_i^r \in \mathrm{SE}(3)$ in the local frame, an additional variable $\boldsymbol{\Delta}^r \in \mathrm{SE}(3)$ is added to the optimization problem to denote each agent's transformation from the local frame to the global frame. The estimation problem for the multi-agent graph is given in \eqref{multi_cost_funcs} as


{\small 
\begin{align}
    \mathrm{X}_{t}^{*} &= \underset {\mathrm{X}, \Delta}{\text{arg min}}  
    \Bigg\{
        \sum_{r=0}^{R-1} \Bigg( 
        \underbrace{ \left\|P(\mathbf{x}_0^r,\mathbf{p}^r)\right\| ^2_{\Sigma_{P}}}_{\text{Pose Prior}}
        + \underbrace{ \left\|A(\boldsymbol{\Delta}^r,\mathbf{a}^r)\right\| ^2_{\Sigma_{A}}}_{\text{Anchor Prior}}
    \nonumber \\
    &  \quad
    + \sum_{i=0}^{M-1}  
    \Bigg(
    \underbrace{\left\|G(\mathbf{x}_{i-1}^0,\mathbf{x}_{i}^0,\mathbf{u}_{i}^0)\right\|_{\Sigma_{G_i}}^{2} }_{\text{Odometry (Computing Agent)}} 
    \Bigg)
    \nonumber \\
    &  \quad
    + \sum_{r=1}^{R-1}\sum_{i=0}^{M_r-1}  
    \Bigg(
    \underbrace{\left\|G(\mathbf{x}_{i-1}^{r},\mathbf{x}_{i}^{r},\mathbf{z}_{i}^{r})\right\|_{\Sigma_{z_i^r}}^{2} }_{\text{Odometry (Remote Agents)}} 
    \Bigg)
        \nonumber \\
        & \quad
            + \sum_{i=0}^{M_r}  
            \Bigg(
            \underbrace{\left\|\Xi({\bf x}_{i}^{r},\boldsymbol{\Delta}^{r},\boldsymbol{\xi}_i^r)\right\|_{\Sigma_{\Xi_i}}^{2} }_{\text{Orientation}}
            + \underbrace{\left\|D({\bf x}_{i}^{r},\boldsymbol{\Delta}^{r}
            ,d_i^r)\right\|_{\Sigma_{D_i}}^{2}}_{\text{Depth}}
            \Bigg)
    \Bigg) 
    \nonumber \\
    &  \quad
    + \sum_{j=1}^{N} \underbrace{
    \left\|C({\bf x}_{i_j}^{0},
            {\bf x}_{i_j^{\prime}}^{r_j^{\prime}},
            \boldsymbol{\Delta}^{0}, 
            \boldsymbol{\Delta}^{r_j^\prime},{\boldsymbol{c}}^j)\right\|_{\Sigma_{C_j}}^{2}
    }_{\text{Inter-Robot Bearing}}
    \Bigg\} 
    \label{multi_cost_funcs}
\end{align}
}

The factors in the multi-agent graph are defined as follows:

\begin{figure*}
  \centering
  \includegraphics[width= \textwidth]{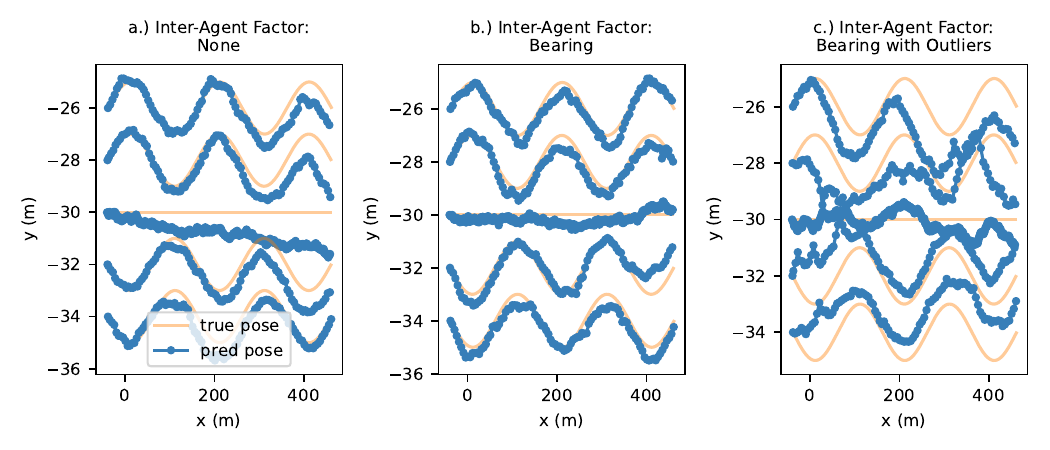}
  \caption{Simulation results implementing factor-graph-based localization using BEDD. For each experiment, the multi-agent factor graph contains the priors, odometry, orientation, and depth factors for each agent. The type of inter-agent factor varies, with a.) no inter-agent factors, b.) azimuth elevation inter-agent factors with no outliers, and c.) azimuth elevation inter-agent factors with 5~percent outliers.}
  \label{results}
\end{figure*}

\subsubsection{Pose and anchor prior}
Each agent is initialized with a unary pose prior factor as described in Section \ref{sec:pose_prior}, located at the agent's local origin. To convert from an agent's local origin to the global frame, each agent is also initialized with an anchor node variable $\boldsymbol{\Delta}^r \in \mathrm{SE}(3)$ and anchor prior $\mathbf{a}^r\in \mathrm{SE}(3)$~\cite{kimAnchorNode2010}. Converting an agent's pose from the local frame $\ell$ to global frame $w$ using an anchor node can be denoted by 
\begin{equation}
    {}^{w}\mathbf{x}_i^{r}=(\boldsymbol{\Delta}^r)({}^{\ell}\mathbf{x}_i^{r})
    \label{anchor}
\end{equation}

The anchor prior error can be calculated in the $\mathrm{SE}(3)$ group space using $A(\boldsymbol{\Delta}^r,\mathbf{a}^r)=E(\hat{\mathbf{a}}^r,\mathbf{a}^r)$, where $\hat{\mathbf{a}}^r=\boldsymbol{\Delta}^r$.

\subsubsection{Odometry} The odometry error for both the computing and neighboring agents is computed using the function $G$ defined in \ref{sec:dead_reck_single}. To add its own odometry factor, the computing agent uses the measured dead reckoning odometry $\mathbf{u}_i$, computed using \eqref{odom}. In contrast, the computing agent adds odometry for neighbors using the summarized transformation $\mathbf{z}_i^r$ according to methods found in Section \ref{sec:origin-state}.

\subsubsection{Orientation and depth} 

Equations \( \Xi \) and \( D \) follow the definitions in Sections \ref{sec:orientation} and \ref{sec:depth}. Before extracting orientation and depth components from the predicted poses in the multi-agent graph, the poses are transformed to the global frame using \eqref{anchor}. This ensures coordinate frame consistency before computing the error with global orientation and depth measurements, \( \boldsymbol{\xi}_i^r \) and \( d_i^r \), for each robot \( r \).

\subsubsection{Inter-robot bearing}
The inter-robot bearing factor partially constrains the relative pose between the computing agent and neighboring agents, where a neighbor corresponding to a specific encounter is denoted as $r_j^{\prime} \in \{1, \dots,R-1\}$. The measured azimuth $\alpha$ and elevation $\epsilon$ are computed by the receiving acoustic modem using a phased hydrophone array. The encounter is represented as a vector $\mathbf{c}^j=[\alpha, \epsilon]$. The function $C$ computes the error between $\mathbf{c}^j$ and the predicted inter-agent bearing $\hat{\mathbf{c}}^j=[\hat{\alpha}, \hat{\epsilon}]$  by first converting the predicted pose of each agent to the global frame using anchor nodes $\boldsymbol{\Delta}^{0}$ and $\boldsymbol{\Delta}^{r_j^\prime}$ according to \eqref{anchor}. Next, each predicted angle is computed by converting the translational difference $[\delta_x, \delta_y, \delta_z]$ of the two poses in the global frame to spherical coordinates as follows:
\begin{align}
    r &= \sqrt{\delta_x^2 + \delta_y^2 + \delta_z^2}, \nonumber \\
    \hat\alpha &= \tan^{-1} \left(\frac{\delta_y}{\delta_x} \right),  \\
    \hat\epsilon &= \cos^{-1} \left(\frac{\delta_z}{r} \right). \nonumber
\end{align}

Instead of using circular differencing as done in~\cite{sekimoriBEDD2024} to calculate azimuth and elevation error, the error is calculated in the $\mathrm{SO}(2)$ group space. First, the azimuth and elevation angles are assumed to be uncorrelated. We use $\beta$ to represent an angle, which may correspond to the predicted or measured azimuth ($\hat{\alpha}, \alpha$) or elevation ($\hat{\epsilon}, \epsilon$). Each measured and predicted angle is converted to the $\mathrm{SO}(2)$ space using

\begin{align*}
    R_{\beta} &= 
    \begin{bmatrix}
        \cos \beta 
        &-\sin \beta \\
        \sin \beta & \cos \beta
    \end{bmatrix} \\ 
\end{align*}

The difference between two rotation matrices in SO(2) space can be calculated using
\begin{align}
    R_{{\beta_{error}}} = R_{\hat\beta}^TR_{\beta},
\end{align}
where $R_{\hat\beta}$ is derived from the predicted angle $\hat \beta$ and $R_{\beta}$ is derived from the observed angle $\beta$. This error is then transformed to the algebra space using the logarithmic map. 

In summary, the error function for this factor can be defined as follows:
\begin{align}
    C({\bf x}_{i_{j}}^{r_j},
            {\bf x}_{i_{j}^{\prime}}^{r_{j}^{\prime}},
            &\boldsymbol{\Delta}^{r_j}, 
            \boldsymbol{\Delta}^{r_j^\prime},{\boldsymbol{c}}^j) = \nonumber \\ 
       &\begin{bmatrix}
        \ln(R_{\hat\alpha}({\bf x}_{i_{j}}^{r_j},
            {\bf x}_{i_{j}^{\prime}}^{r_{j}^{\prime}},
            \boldsymbol{\Delta}^{r_j}, 
            \boldsymbol{\Delta}^{r_j^\prime})^TR_{\alpha}(\boldsymbol{c}^j)) \\
        \ln(R_{\hat\epsilon}({\bf x}_{i_{j}}^{r_j},
            {\bf x}_{i_{j}^{\prime}}^{r_{j}^{\prime}},
            \boldsymbol{\Delta}^{r_j}, 
            \boldsymbol{\Delta}^{r_j^\prime})^TR_{\epsilon}(\boldsymbol{c}^j))
    \end{bmatrix}.
\end{align}



Throughout the mission, each agent updates its own nodes in both its multi-agent and single-agent graph using odometry and unary factors drived from it's own onboard sensors. Upon receiving a message, an agent also adds odometry factors obtained from decomposing the remote chain summaries broadcast from the remote vehicle, any transmitted unary factors, and the inter-agent factor to its multi-agent factor graph enabling corrections to (1) the estimated relative transformation between agents in the multi-robot team and (2) the computing agent's own estimated pose.

\section{Experiments}

The multi-agent localization factor-graph framework using BEDD has been implemented and tested in the HoloOcean underwater simulator~\cite{potokarHoloOcean2022} with a fleet of five underwater torpedo vehicles. The Georgia Tech Smoothing and Mapping (GTSAM) library~\cite{gtsam} was used for factor-graph evaluation. 

The multi-agent factor-graph was evaluated first with no inter-agent constraints, then with bearing (azimuth and elevation) inter-agent contraints. As shown in Fig. \ref{results}a, when only the broadcasting agent's odometry, orientation, and depth are incorporated into the computing agent's multi-agent factor graph, significant drift accumulates over time. In contrast, Fig. \ref{results}b illustrates that introducing inter-agent constraints effectively mitigates this drift, even in the absence of landmark observations. While the absence of GPS and landmark measurements can lead to long-term drift in the overall system in the global frame, regular inter-agent measurements constrain the relative positional error between agents.

To analyze system performance under simulated multi-path effects, we conducted an experiment where outliers were added to the set of inter-agent bearing measurements. The simulated bearing noise was modeled as additive zero mean Gaussian noise with a standard deviation of 10~degrees for both azimuth and elevation. Approximately 5~percent of the measurements were designated as outliers, with their values sampled from a uniform distribution between 40~degrees and 120~degrees. Fig.~\ref{results}c. demonstrates how localization estimates degrade in the presence of outliers, motivating the need for future research in bearing-related consistency metrics for outlier rejection.

We note that for this initial implementation, the covariances $\Sigma _{G_i}$ associated with odometry factors $\mathbf{z}_{i}$ were not computed using proper decomposition methods, but were instead tuned to reflect the compounding uncertainty for the approximate local chain. In our next iteration, we plan to correctly recover the covariances using the uncertainty propagation methods presented by~\cite{mangelsonUncertainty2020,barfootUncertainty2014}. 

\begin{figure}
  \centering
    \includegraphics[width=0.47\textwidth]{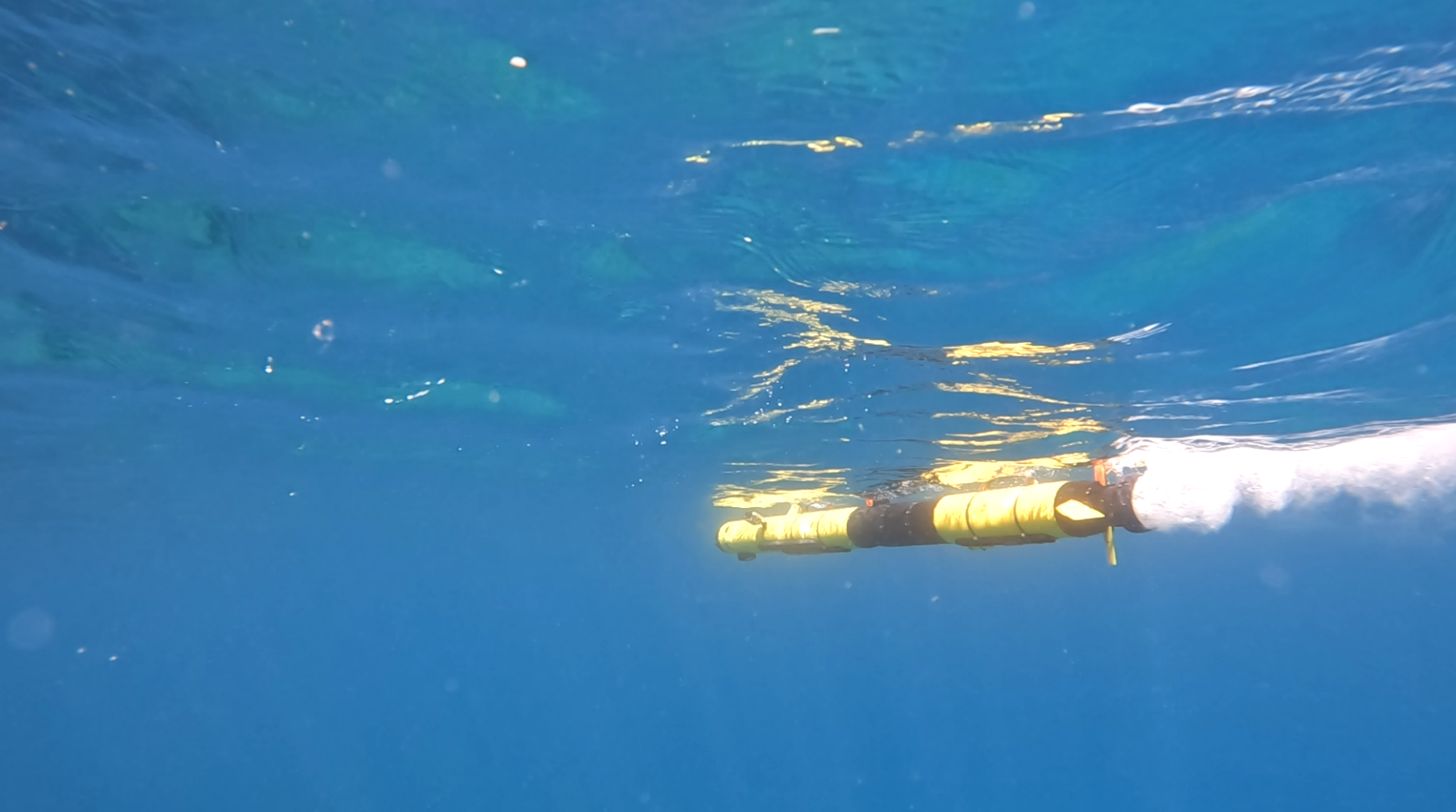}
  \includegraphics[width=0.47\textwidth]{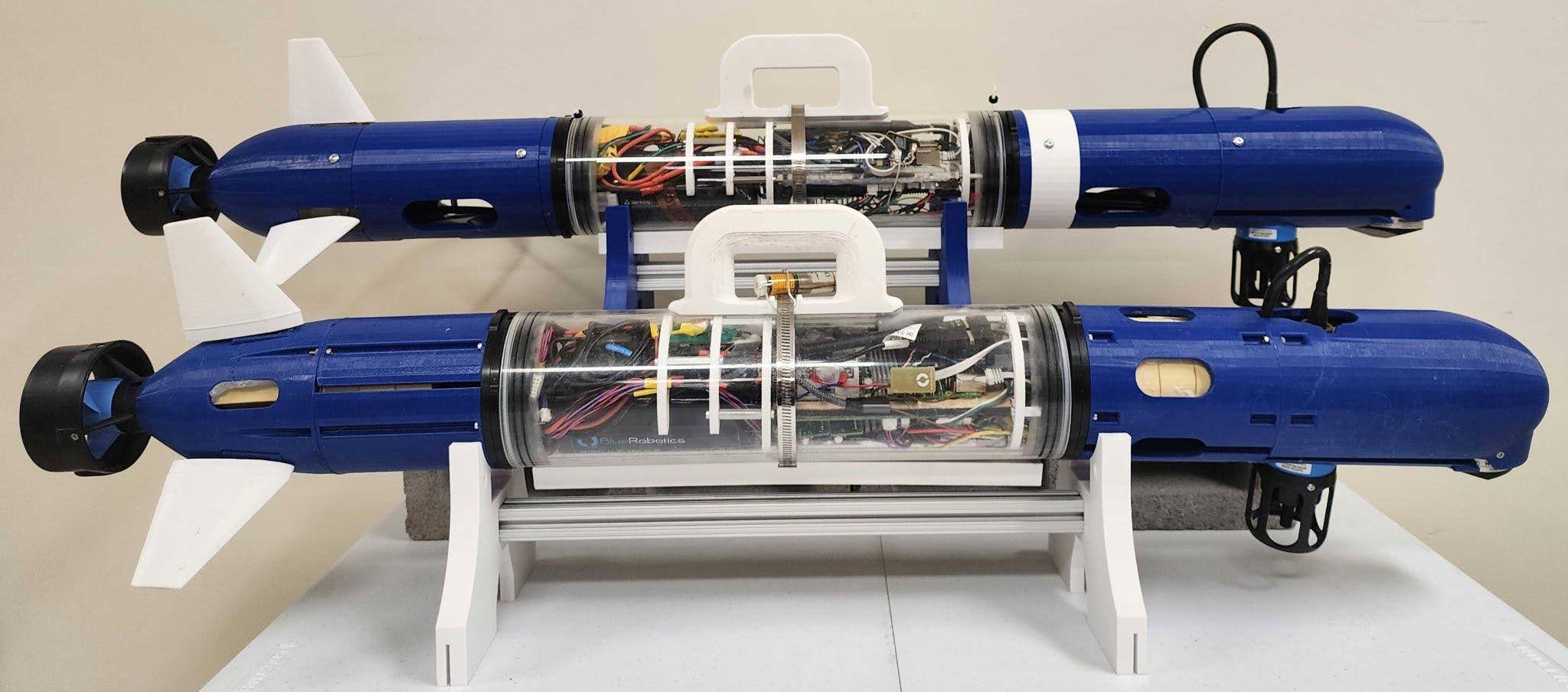}
  \caption{Top: BYU Field Robotic Systems Lab Iver3 autonomous underwater vehicle (AUV) conducting an underwater mission. Bottom: Two CougUV low-cost underwater vehicles. We plan to utilize a fleet composed of 2 IVER3 AUVs and multiple CougUVs to evaluate the proposed techniques in the field.}
  \label{cougUV}
\end{figure}

In addition, to bridge the gap between simulation and real-world deployment, we plan to conduct field testing experiments with two IVER3 unmanned underwater vehicles (UUVs) and a fleet of low-cost UUVS currently under construction in our (Fig. \ref{cougUV}). We will use this fleet to validate our localization framework via in-water experiments. Initial single-agent odometry pose graph tests have been conducted on the Coug-UVs pictured. Multi-agent field tests are planned for the near future to assess performance under real-world conditions.

\section{Summary}

In this work, we presented a factor-graph-based framework for multi-agent localization using BEDD constraints. We effectively leverage inter-agent azimuth, elevation, and depth measurements to correct positional drift of a simulated fleet. Our inverted-USBL approach is computationally scalable and robust compared to TW-ToF and particle filter-based BEDD methods.

Future work will focus on improving the resilience of our framework to measurement errors by exploring outlier rejection techniques. Specifically, we aim to incorporate consistent set maximization methods~\cite{mangelsonPCM} and azimuth-elevation consistency metrics to mitigate the effects of multipath-induced errors. Additionally, we plan to conduct fully underwater multi-agent hardware experiments, where we anticipate reduced multipath effects when operating at greater depths. In the longer term, we envision integrating terrain-aided navigation techniques to further enhance localization accuracy in complex underwater environments. These advancements aim to contribute towards scalable and cost-effective underwater multi-agent localization solutions.

\section{Acknowledgement}
We used ChatGPT to assist with grammar and wording refinement.


\bibliographystyle{style/IEEEtranN}
\bibliography{refs}


\end{document}